# CGM: Chaos-Guided Graph Attention and Expert Routing for Autism Stereotypy Detection

Hafsa Aslam, Yue Li, Saba Aslam, Abdur Rasool, Qiao Pan, Muhammad Rizwan, and Ovidiu Bagdasar

***Abstract*—Skeleton graph attention for autism video analysis weights joints by content alone, giving an erratic wrist the same standing as a still one. Stereotypy, however, is defined by how unpredictably movement unfolds rather than how much of it occurs: in our cohort, every amplitude-based kinematic descriptor falls in stereotyped clips, while windowed Sample Entropy of angular velocity alone rises (Hedges' $g = +1.05$). We compute this irregularity as a per-joint, per-frame tensor that structurally drives computation. The proposed chaos-guided graph attention and mixture-of-experts model (CGM) couples two mechanisms. First, a learned pairwise term added to the graph-attention logit concentrates attention on irregular joints and re-ranks the remaining edge weight toward jointly irregular pairs; its unconstrained coefficient neither vanishes nor inverts, and zeroing the tensor realigns attention in $94$ % of clips. Second, a gate conditioned on the same signal routes each frame to two of four experts, whose weights track frame-level irregularity. Under subject-grouped five-fold cross-validation on $116$ clips, CGM reaches $0.930$ balanced accuracy, exceeding the graph-attention baseline by $+0.187$; the component ordering replicates on the public MMASD dataset, and on held-out Panoptic Studio sessions, CGM keeps $0.86$ specificity without retraining, against at most $0.35$ for chaos-free baselines. In ablation, routing without irregularity yields no reliable gain, whereas chaos modulation alone contributes $+0.132$. A severity index from the same tensor reaches $0.87$ split-half reliability, separates the cohorts with Hedges' $g = +1.08$, and stays consistent within children across therapy recordings, supporting video-only monitoring of stereotypy during intervention.**

## I. Introduction

AUTISM spectrum disorder (ASD) affects 1 in 31 U.S. 8-year-olds, yet diagnosis lags behind clinical identification [1], [2]. Motor atypicalities often precede social-communication symptoms, making movement an early signal [3], [4], one whose unpredictability, more than its magnitude, separates autistic from typical movement in rehabilitation settings. Stereotyped movements such as hand flapping and body rocking are core diagnostic criteria [5] and clinically significant [6], and tracking them across therapy sessions is central to rehabilitation monitoring progress in behavioural intervention [7], yet they remain hand-assessed [8]. Human-in-the-loop home-video methods balance human and AI features [9], but automated irregularity-conditioned stereotypy detection remains limited. Because the human body forms a graph of joints linked by limbs, graph attention networks (GATs) and related graph neural networks have become the standard basis for automated movement analysis [10]–[14], and graph learning now underpins neurodevelopmental assessment from brain connectivity [15]–[17], audio-visual behaviour, and skeleton motion [18]–[21].

Despite this promise, three weaknesses limit the rehabilitation value of current methods: signal-blind architecture, descriptor impoverishment, and validation fragmentation. First, standard graph attention is signal-blind: it computes edge weights from node content alone, treating an erratic wrist identically to a still one because no mechanism lets temporal irregularity shape the computation [22]. Skeleton-specific extensions refine that comparison through spatio-temporal aggregation and learned edge weighting [11], [12], [23], [24], without changing what it is computed from. Where an external signal is admitted, it is admitted as content: Zahan *et al.* [25] embed a pairwise joint-angle matrix into the input skeleton of a graph convolutional encoder, enriching node content rather than edge weighting. Sparse mixture-of-experts (MoE) routing allocates capacity across domains [26]–[28] and now reaches skeleton action analysis, where gates follow motion velocity [29] or self-supervised spatio-temporal features [30]; to our knowledge, none conditions routing on a measured, interpretable irregularity signal.

Second, conventional kinematic features measure movement magnitude rather than predictability. Sample Entropy [31], a proven measure of unpredictability, has quantified motor irregularity across human-movement science [32], and entropy measures have distinguished autistic from typically developing children via arm-movement velocity [33], postural sway [34], and head-movement complexity [35], and have classified self-injurious behaviour [36]; yet such measures enter only as clip- or episode-level summary features, discarding the frame-to-frame structure that distinguishes stereotyped from typical movement. They also inherit the unverified limitations of home-video pose estimation [37], [38]. Lastly, validation fragmentation compounds both: over half of thirteen home-video ASD models were never cross-validated [39], a pattern echoed across related pipelines [40], and motor-focused ASD

This work was supported by the National Key Research and Development Program of China under Grant 2024YFE0102100. The authors declare no competing interests. *(Corresponding authors: Yue Li and Abdur Rasool.)*

Hafsa Aslam, Yue Li, and Qiao Pan are with the School of Information and Intelligent Science, Donghua University, Shanghai 200051, China (e-mail: 324091@mail.dhu.edu.cn; frankyueli@dhu.edu.cn; panqiao@dhu.edu.cn).

Saba Aslam is with the Guangdong Provincial Key Laboratory of Multimodality Non-Invasive Brain-Computer Interfaces, Shenzhen Institutes of Advanced Technology, Chinese Academy of Sciences, Shenzhen 518055, China (e-mail: aslam@siat.ac.cn).

Abdur Rasool is with the Department of Information and Computer Sciences, University of Hawaii at Manoa, Honolulu, HI 96822, USA (e-mail: abdur@hawaii.edu).

Muhammad Rizwan and Ovidiu Bagdasar are with the College of Science and Engineering, University of Derby, Derby, UK (e-mail: m.rizwan@derby.ac.uk; o.bagdasar@derby.ac.uk).

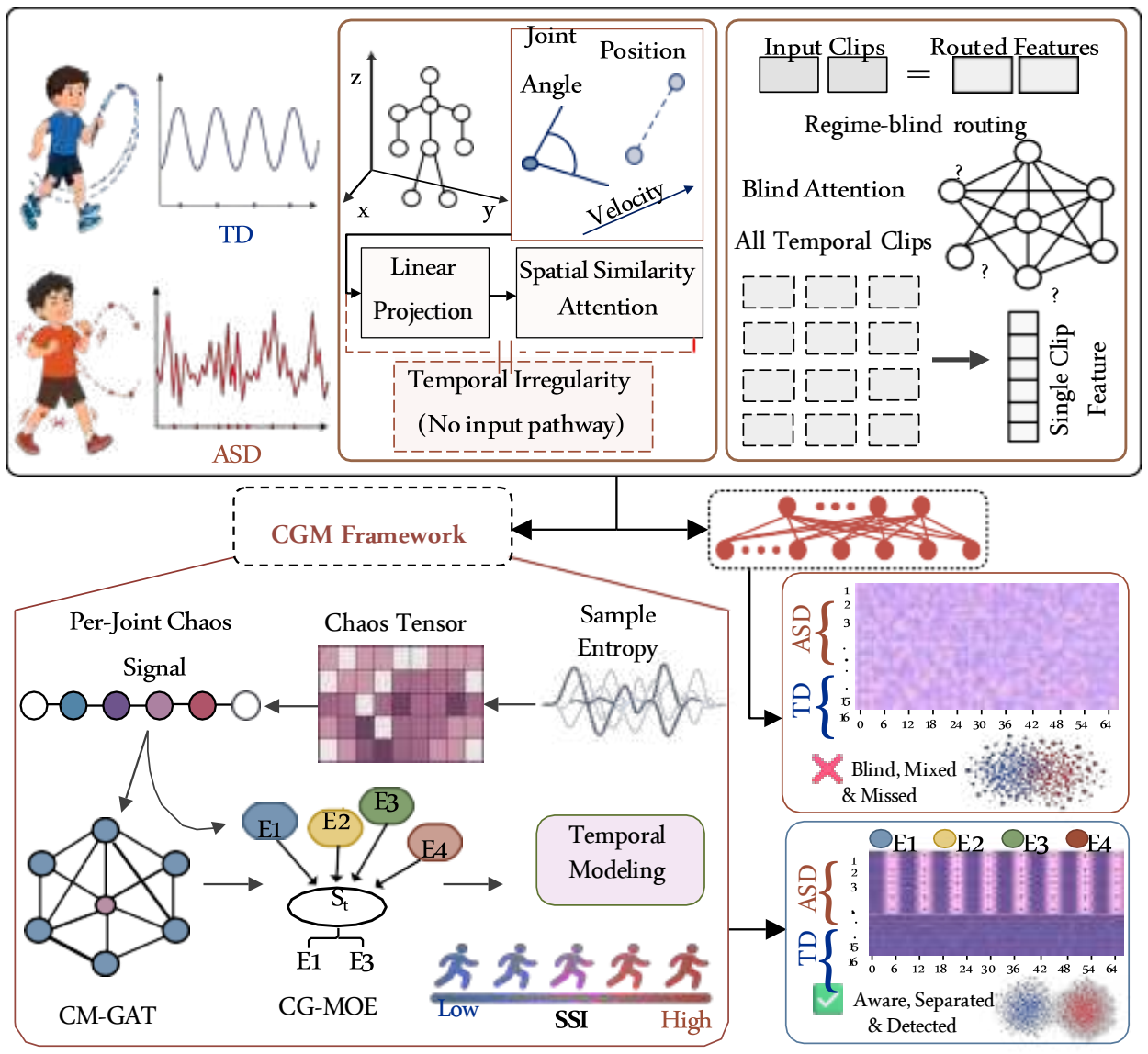

Fig. 1. Temporal irregularity has no pathway in conventional skeleton pipelines: attention follows spatial similarity alone and routing treats calm and chaotic frames alike. CGM supplies this pathway by deriving a per-joint Sample Entropy tensor that modulates inter-joint attention and guides sparse expert selection per frame.

work [8], [41]–[43], reports accuracy without testing whether it arises from the intended movement signal or from acquisition shortcuts, a failure mode documented in medical imaging [44], or whether the design holds on an independent cohort (Fig. 1).

To address these gaps, we argue that video-based motor screening must be irregularity-conditioned at the structural level, not the feature level. We introduce CGM, a chaos-guided graph attention and mixture-of-experts model built on a windowed Sample Entropy tensor over a seven-node body graph. The tensor enters the network at two points: (i) the graph-attention logit, so that attention concentrates on erratic joints and prefers edges whose endpoints are jointly erratic; and (ii) a sparse expert gate, so that capacity follows the motion regime of each frame. A shared Transformer encoder [45] then aggregates the routed sequence over time. We validate both mechanisms by intervention rather than accuracy: zeroing and restoring the tensor attributes attention shifts to the modulation term, blinding the gate removes its routing pattern, and an ablation admitting the same signal as an input feature isolates the effect of injecting it at the attention logit. CGM achieves high balanced accuracy under cross-validation grouped with subjects; its component ordering is generalized to the MMASD dataset [7], recorded during play therapy, and its decision boundaries are transferred to holdout recordings without retraining. The core contributions are:

1) We propose an attention mechanism that overcomes the "blind attention" of spatial-only networks: a windowed Sample Entropy tensor injected into the attention logits steers focus onto erratic joints and toward jointly erratic pairs, rather than entering as a passive input feature.
2) We devise a sparse MoE gate conditioned on the same signal, allocating capacity per frame by motion regime, so that erratic stereotypies receive dedicated experts instead of a uniform share.
3) We establish windowed Sample Entropy as an irreducible motor descriptor that, unlike magnitude-based kinematics, rises in stereotypy, and derive from it a continuous Stereotypy Severity Index (SSI) that supplies a graded score alongside binary detection and remains consistent within children across therapy recordings.

## II. Methodology

### A. *Problem Formulation*

*Notation.* A clip $V = \{I_t\}_{t=1}^{T}$ of $T$ frames carries a label $y \in \{0, 1\}$ (TD comparison, ASD stereotypy). Pose extraction maps $V$ to a spatio-temporal graph $G = (J, A, \mathbf{X})$ with $|J| = 7$ anatomical nodes, fixed adjacency $A \in \{0, 1\}^{7\times7}$, and node features $\mathbf{X} \in \mathbb{R}^{T\times7\times4}$. From the same clip, and no further sensor, we derive an irregularity tensor $\mathbf{C} \in \mathbb{R}^{T\times7}$ with entries $c_{t,j}$, one per frame and joint.

*Objective.* Learn a classifier $f_\theta : G \to [0, 1]$ with parameters $\vartheta$ that maximises the expected log-likelihood over the clip distribution $D$,

$$\vartheta^\star = \arg\max_{\theta} \ \mathbb{E}_{(V,y)\sim D}\left[\log p_\theta(y \mid G)\right], \qquad (1)$$

from camera video alone, without markers or wearables.

*Constraint.* Graph attention scores edge $(i, j)$ from node content alone, $e_{ij} = g_\phi(\mathbf{x}_i, \mathbf{x}_j)$ with scoring function $g_\phi$ [22], so two clips of matching content receive matching attention however predictably that content evolves. Stereotypy is temporal organisation rather than magnitude, so an admissible score must also see the irregularity of both joints,

$$e_{ij} = g_\phi(\mathbf{x}_i, \mathbf{x}_j, c_i, c_j). \qquad (2)$$

CGM satisfies Eq. (2) by deriving $\mathbf{C}$ from the clip itself (Fig. 2) and injecting it at two points of the network (Fig. 3): the attention logit, which decides which edges attention prefers, and the expert gate, which decides how much capacity a frame receives. Supplying $\mathbf{C}$ as a node feature forms no pairwise term and is retained only as an ablation baseline.

### B. *Dataset and Preprocessing*

The cohort merges SSBD [41] with its extension SSBD+ [42], which restores coverage lost to platform takedowns and adds 35 clips in the same three categories, giving 86 ASD stereotypy clips from rehabilitation and home settings; the 30 typically developing (TD) comparison clips come from one sequence (170915_toddler5) of the toddler subset of the public CMU Panoptic Studio dataset [46], [47], which features a single two-year-old child at play with caregivers under fixed camera, lighting, and background and has served as a validation set for automated caregiver–child interaction analysis [48]; its unseen sequences form a held-out test (Section III-E5). Each ASD clip shows a different child, identified by its source video, which serves as the subject identifier for grouping. All four datasets used here are public and de-identified, and no new human-subject data were collected, so no institutional ethics review was required. Each clip then passes through five sequential steps.

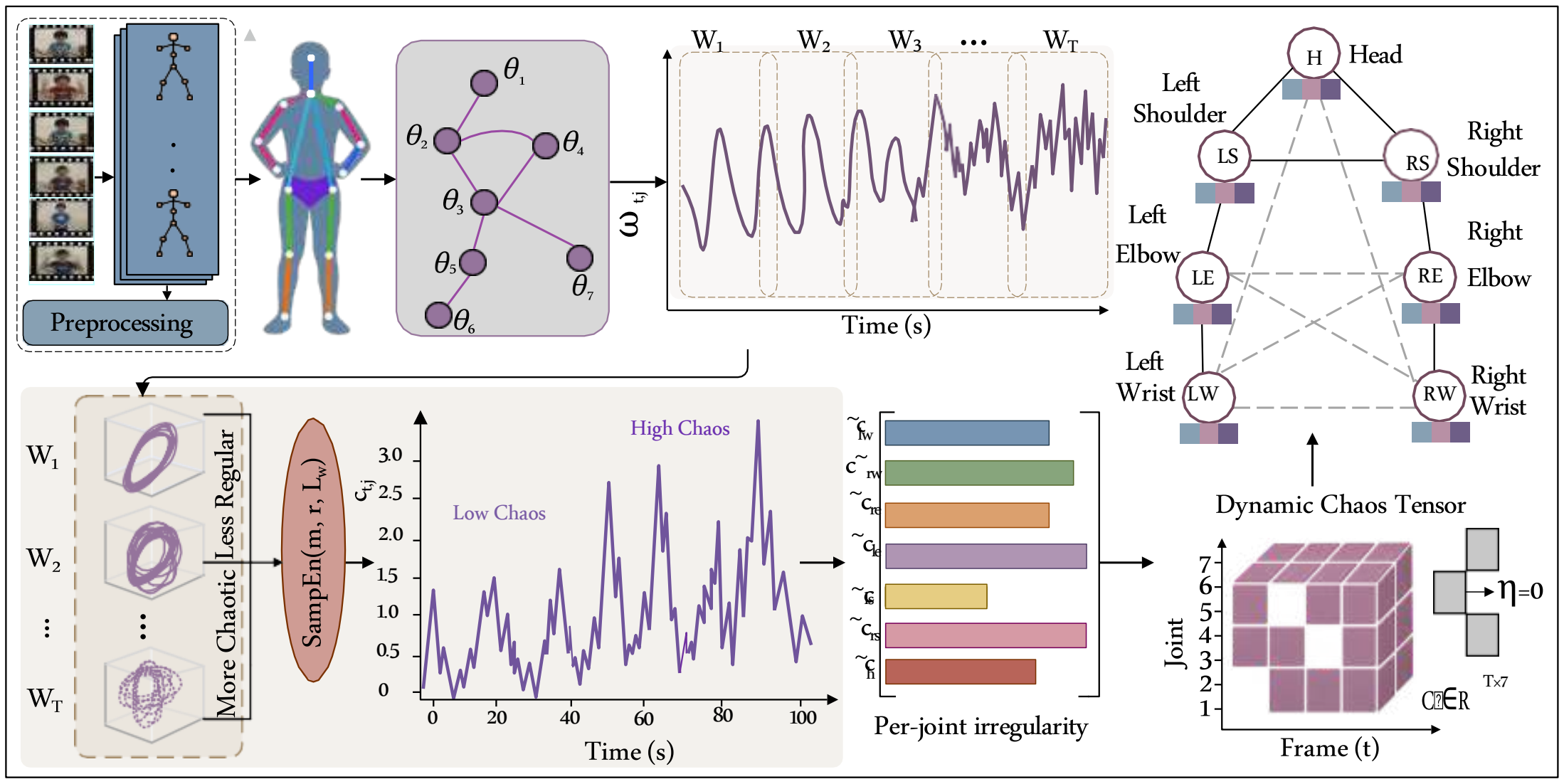

Fig. 2. Chaos Formation. A seven-node anatomical graph is converted to timestamp-aware angular velocity $\omega_{t,j}$ , isolating movement dynamics from static posture. Sample Entropy is evaluated on a stride-one window to quantify moment-to-moment unpredictability independent of magnitude; insets order three windows from regular to irregular. Stacking these estimates yields the dynamic chaos tensor that drives attention and routing in CGM.

1) *Pose extraction.* MediaPipe Pose Landmarker [49], on BlazePose [50], extracts 33 landmarks with visibility $v_{t,a}$ for landmark $a$; seven upper-body nodes are kept,

$$J = \{\text{L-W, R-W, L-E, R-E, L-S, R-S, H}\}, \tag{3}$$

the left (L) and right (R) wrist (W), elbow (E), and shoulder (S), and the head (H). Dense facial landmarks are excluded because usable face detection reached only 17 % of frames overall and 1 % in the comparison recordings.

2) *Visibility masking.* A node is valid if every landmark $a$ in its landmark set $A_j$ clears a 0.5 threshold,

$$\mu_{t,j} = \mathbf{1}\left[\min_{a\in A_j} v_{t,a} \ge 0.5\right], \tag{4}$$

and invalid nodes are masked rather than zero-filled.

3) *Normalisation.* Position $\bar{\mathbf{p}}_{t,j}$ is anchored on the shoulder midpoint and scaled by shoulder width.

4) *Velocity and node features.* Velocity uses frame timestamps $\tau_t$ over valid consecutive pairs,

$$\mathbf{u}_{t,j} = \frac{\bar{\mathbf{p}}_{t,j} - \bar{\mathbf{p}}_{t-1,j}}{\tau_t - \tau_{t-1}}, \tag{5}$$

giving node feature $\mathbf{x}_{t,j} = [\bar{x}_{t,j}, \bar{y}_{t,j}, u^x_{t,j}, u^y_{t,j}]^T \in \mathbb{R}^4$.

5) *Resampling and graph.* Sequences are resampled to $T_* = 64$ tokens after masking. The fixed adjacency $A$ (wrist–elbow, elbow–shoulder, shoulder–shoulder, head–shoulder, and self-loops) completes the graph.

For generalisation, the four repetitive-movement activities of the public MMASD release [7] (arm swing, body swing, maracas forward shaking, maracas shaking; 453 clips, 22 children with ASD) use the same nodes, Sample Entropy, and model, with joints read from the released OpenPose skeletons (Supplementary Section S6.1).

## C. Chaos Formation

*1) Angular Kinematics over the Body Graph:* Seven biomechanical angles (Fig. 2) are computed per frame (left/right elbow flexion, left/right shoulder angle, left/right wrist angle, and head–shoulder angle) from the same landmarks and under the same visibility rule as $\mu_{t,j}$, giving one angle $\vartheta_{t,j}$ per node of $J$ . The architecture is built on angular velocity,

$$\omega_{t,j} = \frac{\vartheta_{t,j} - \vartheta_{t-1,j}}{\tau_t - \tau_{t-1}}, \tag{6}$$

rather than raw angle, since irregularity is a property of how motion changes over time, not of static posture: two clips can share a posture yet differ entirely in how they reach it.

*2) Windowed Sample Entropy:* Each clip's angles are placed on a uniform grid at its median frame interval $\Delta$, interpolating linearly only across gaps of at most 1 s, and $\omega_{t,j}$ is taken on this grid; for the head node, $\omega$ is replaced by the speed of the nose relative to the shoulder midpoint, normalised by shoulder width. For node $j$, a window $W_{t,j}$ of $L_w = \max\{16, \mathrm{round}(8\ \mathrm{s}/\Delta)\}$ grid samples, capped at the clip length, centred at frame $t$ and shifted rather than truncated at the clip edges, collects the values in range. Windowed Sample Entropy [31] is computed on this window with embedding dimension $m = 2$ and tolerance $r_j = 0.2\sigma_j$, where $\sigma_j$ is the standard deviation of node $j$'s signal over the whole clip [31], [32]. Writing $U^{(m)}_{t,j}$ and $U^{(m+1)}_{t,j}$ for the counts of length-$m$ and length-$(m+1)$ template matches within tolerance $r$, self-matches excluded,

$$c_{t,j} = \mathrm{SampEn}(W_{t,j}) = -\log\left(\frac{U^{(m+1)}_{t,j}}{U^{(m)}_{t,j}}\right). \tag{7}$$

The estimate is evaluated in every frame with a valid pose, $c_{t,j}$ is a per-frame value rather than a block summary, and values still missing after gap filling are removed from the window rather than imputed. A window with < 16 remaining samples, or with $U^{(m)}_{t,j} = 0$ or $U^{(m+1)}_{t,j} = 0$, is marked invalid by a second indicator $\eta_{t,j} = \mathbf{1}[c_{t,j}$ is estimable], mirroring $\mu_{t,j}$.

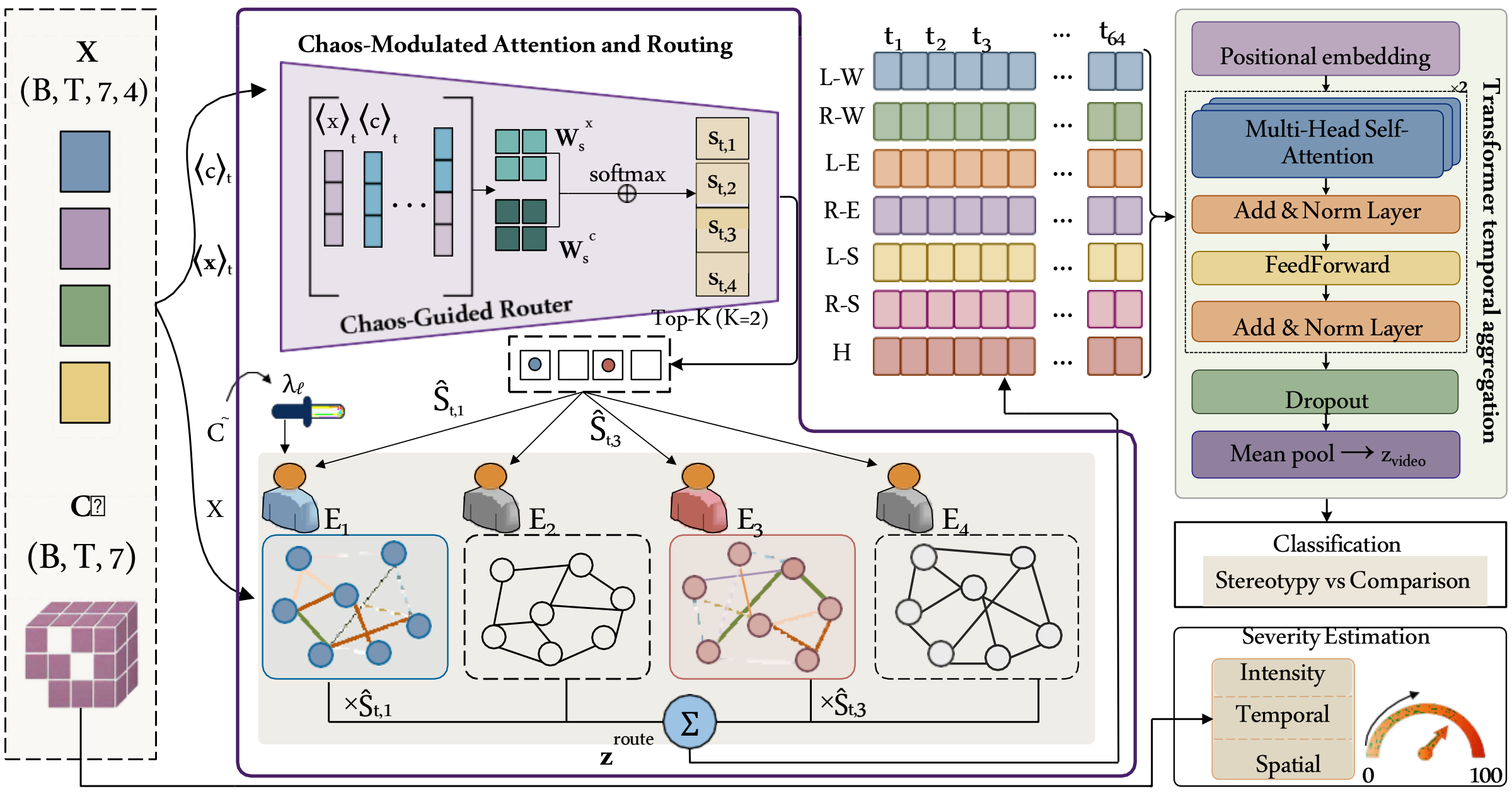

Fig. 3. Proposed CGM framework for video-based motor stereotypy analysis. Inputs are node features $\mathbf{X}$ and the chaos tensor $\tilde{\mathbf{C}}$ of Fig. 2. The chaos-guided router scores four experts from pooled motion $\langle \mathbf{x} \rangle_t$ and pooled chaos $\langle c \rangle_t$ through the weight blocks $\mathbf{W}^x_s$ and $\mathbf{W}^c_s$, keeping the top two. Each expert is a chaos-modulated graph attention network (CM-GAT) whose edge weights are reshaped by the irregularity of both joints, while irregularity never enters node features. A shared Transformer encoder aggregates the routed sequence $\mathbf{z}^{\text{route}}$ for ASD stereotypy vs. TD comparison classification, and the Stereotypy Severity Index (SSI) is computed from $\tilde{\mathbf{C}}$ on a parallel branch.

*3) Tensor Assembly and Per-Clip Scaling:* The tensor $\mathbf{C}$ and its indicator $\eta$ are ordered as $\mathcal{J}$ and resampled with $\mathbf{x}$ onto the common $T_* = 64$ grid, so each value describes irregularity at the corresponding relative position in the clip. Each clip $v$ is then scaled on its own estimable entries,

$$\tilde{c}_{t,j} = \min\left(1, \max\left(0, \frac{c_{t,j} - q^{(v)}_2}{q^{(v)}_{98} - q^{(v)}_2}\right)\right), \tag{8}$$

where $q^{(v)}_2$ and $q^{(v)}_{98}$ are the 2nd and 98th percentiles of clip $v$'s estimable entries, giving the scaled tensor $\tilde{\mathbf{C}}$.

This scaling has three properties the design relies on. First, $\tilde{c} \geq 0$, so a product of two entries is large only when both joints are irregular at that frame, and two quiet joints cannot raise each other's attention. Second, it draws on no other clip, so no normalisation parameter is fitted across clips and nothing can leak between folds. Third, it removes between-clip differences in the absolute level of irregularity, which can reflect recording conditions, so the network acts on where and when irregularity rises within a clip. Percentile bounds keep the scaling robust to isolated estimation outliers.

## D. *Chaos-Modulated Graph Attention (CM-GAT)*

*Motivation.* Stereotypies are expressed by joints acting together: arm flapping drives both wrists and elbows into irregular motion at once, and rocking couples the head with both shoulders. Standard graph attention [22] cannot represent this co-occurrence, because its edge scores see only the instantaneous content of the two joints. CM-GAT supplies the missing pathway by making joint irregularity a pairwise prior on the attention logit, inside every expert of the network.

*Formulation.* Writing $\mathbf{h}^{(\ell-1)}_{t,j}$ for node $j$'s hidden state entering layer $\ell \in \{1, 2\}$ (with $\mathbf{h}^{(0)}_{t,j} = \mathbf{x}_{t,j}$, the kinematic feature of step 4 and nothing further), and $\mathbf{z}^{\ell,k}_{t,j} = \mathbf{W}^{\ell,k}\mathbf{h}^{(\ell-1)}_{t,j}$ its projection by the learnable matrix $\mathbf{W}^{\ell,k}$ of head $k \in \{1, \ldots, 4\}$, standard attention scores edge $(i, j)$ as

$$e^{\ell,k,\text{std}}_{t,ij} = \text{LReLU}_{0.2}\left((\mathbf{a}^{\ell,k}_s)^{\mathrm{T}}\mathbf{z}^{\ell,k}_{t,i} + (\mathbf{a}^{\ell,k}_d)^{\mathrm{T}}\mathbf{z}^{\ell,k}_{t,j}\right), \tag{9}$$

with learnable source and destination vectors $\mathbf{a}^{\ell,k}_s$, $\mathbf{a}^{\ell,k}_d$ and a leaky rectifier of slope 0.2 [22]. CM-GAT adds a pairwise structural term,

$$e^{\ell,k}_{t,ij} = e^{\ell,k,\text{std}}_{t,ij} + \lambda_\ell\, \tilde{c}_{t,i}\tilde{c}_{t,j}, \tag{10}$$

where $\lambda_\ell \in \mathbb{R}$ is a learned scalar per attention layer, eight in total (two in each of the four experts), initialised at 1 and left free of any constraint, so the optimiser may shrink or invert the term. Because $\tilde{c} \geq 0$, an edge is favored only when both endpoints are irregular in that frame, relative to their own clip, which no per-node feature expresses. The attention coefficient,

$$\alpha^{\ell,k}_{t,ij} = \text{softmax}_j\left(e^{\ell,k}_{t,ij}\right) \tag{11}$$

taken over pairs with $A_{ij}\, \mu_{t,i}\eta_{t,i}\, \mu_{t,j}\eta_{t,j} = 1$, so both the visibility rule of Eq. (4) and the estimability indicator $\eta$ bind here and an inestimable window cannot contribute to any edge weight. The head output is $\mathbf{o}^{\ell,k}_{t,i} = \sum_j \alpha^{\ell,k}_{t,ij}\mathbf{z}^{\ell,k}_{t,j}$; the four head outputs are concatenated and passed through dropout and an ELU nonlinearity to give the node state $\mathbf{H}^{(\ell)}_t \in \mathbb{R}^{7\times h_a}$, with per-node width $h_a = 128$.

*What the term changes.* Because softmax normalises over a fixed edge set, the modulation cannot add attention mass to the graph; it redistributes existing mass onto irregular joints and toward jointly irregular pairs. For node $i$, the self-logit gains $\lambda_\ell \tilde{c}^2_{t,i}$ against $\lambda_\ell \tilde{c}_{t,i}\tilde{c}_{t,j}$ for neighbour $j$, so a joint that is more irregular than its neighbours sharpens onto itself, while the weight left on skeletal edges is re-ranked by the product of their endpoints' irregularity.

### E. *Chaos-Guided Sparse Expert Routing (CG-MoE)*

A single attention mechanism must represent every motion regime a clip contains, from near-stillness to highly irregular movement, with one shared parameter set. CG-MoE (Fig. 3) instead instantiates $E = 4$ experts, denoted E1–E4, each a stack of two CM-GAT layers ("×2" in Fig. 3) followed by a feed-forward projection, producing expert output $\boldsymbol{\phi}^{(e)}_{t,j}$ for expert $e \in \{1,\ldots,E\}$. The router receives whole-body summaries of content and irregularity, each averaged over the seven nodes of $\mathcal{J}$:

$$\langle \mathbf{x} \rangle_t = \frac{1}{|\mathcal{J}|}\sum_{j\in\mathcal{J}} \mathbf{x}_{t,j}, \qquad \langle c \rangle_t = \frac{1}{|\mathcal{J}|}\sum_{j\in\mathcal{J}} \tilde{c}_{t,j}, \tag{12}$$

and produces a softmax gate score over experts [26],

$$\mathbf{s}_t = \mathrm{softmax}(\mathbf{W}^x_s \langle \mathbf{x} \rangle_t + \mathbf{W}^c_s \langle c \rangle_t + \mathbf{b}_s) \in \mathbb{R}^E, \tag{13}$$

where the learnable blocks $\mathbf{W}^x_s$ and $\mathbf{W}^c_s$ weight pooled motion and pooled chaos and $\mathbf{b}_s$ is a bias. Both summaries enter the gate by design: conditioning on irregularity alone would blind the router to the motion it is allocating capacity for, while conditioning on content alone would reproduce the regime-blind allocation this mechanism replaces. Following the sparsely gated formulation [26], only the top $K = 2$ entries of $\mathbf{s}_t$ are retained and renormalised to $\hat{s}_{t,e}$ (the "Top-$K$" block of Fig. 3), and the routed representation is

$$\mathbf{z}^{\mathrm{route}}_{t,j} = \sum_{e=1}^{E} \hat{s}_{t,e}\, \boldsymbol{\phi}^{(e)}_{t,j}. \tag{14}$$

We set $K = 2$ by comparing $K \in \{1, 2, 3, 4\}$ on training folds only: $K = 1$ collapsed toward single-expert behaviour, and $K > 2$ added computation without gain.

Sparse routing admits a failure mode in which the gate concentrates on a subset of experts and the remainder receive no gradient. To discourage this collapse, the dense importance of expert $e$ over a batch of $B$ clips, $q_e = (BT_*)^{-1}\sum_{v,t} s_{v,t,e}$, enters a load-balancing penalty in squared-coefficient-of-variation form [26],

$$\mathcal{L}_{\mathrm{bal}} = \frac{\mathrm{Var}(q_1,\ldots,q_E)}{\left(\frac{1}{E}\sum_e q_e\right)^2 + \epsilon}, \tag{15}$$

with $\epsilon$ a small constant.

*Temporal aggregation and training.* For each node $j$, the routed states $\mathbf{z}^{\mathrm{route}}_{t,j}$ over the $T_*$ frames form one temporal sequence. A two-layer Transformer encoder [45] with a learned positional embedding, each layer pairing multi-head self-attention and a feed-forward block under residual Add & Norm (Fig. 3), encodes every node's sequence with shared weights ($d_{\mathrm{model}} = 128$, four heads, dropout 0.3). Mean pooling over time and nodes gives $\mathbf{z}_{\mathrm{video}} \in \mathbb{R}^{128}$, which a two-layer head maps to $\hat{y}$. Training minimises class-weighted cross-entropy

**Algorithm 1** CGM: Irregularity-Driven Attention Modulation and Expert Selection

**Input:** Clips $\{(V, y)\}$, adjacency $A$, $m = 2$, $T_* = 64$, $E = 4$, $K = 2$, $w_{\mathrm{bal}}$
**Output:** Weights $\vartheta$ with $\{\lambda_\ell\}$; per clip $\hat{y}$ and $\mathrm{SSI}_v$
*Chaos formation (once per clip)*
**for** each clip $v$ **do**
  Extract pose; obtain $\boldsymbol{\mu}_{t,j}$ and $\mathbf{x}_{t,j}$ — Eqs. (3)–(5)
  Compute angular velocity $\boldsymbol{\omega}_{t,j}$ — Eq. (6)
  **for** each node $j$ and valid frame $t$ **do**
    **if** window $W_{t,j}$ is estimable **then**
      $c_{t,j} \leftarrow \mathrm{SampEn}(W_{t,j})$; $\eta_{t,j} \leftarrow 1$ — Eq. (7)
    **else**
      $\eta_{t,j} \leftarrow 0$
  Resample to $T_*$; scale to $\tilde{\mathbf{C}}$ — Eq. (8)
*Training (mini-batches of $B$ clips)*
**repeat**
  **for** each clip and frame $t$ **do**
    $\mathbf{s}_t \leftarrow$ gate on $\langle \mathbf{x} \rangle_t$, $\langle c \rangle_t$ — Eqs. (12)–(13)
    $\mathcal{K}_t \leftarrow$ top-$K$ of $\mathbf{s}_t$; renormalise to $\hat{s}_{t,e}$
    **for** each expert $e \in \mathcal{K}_t$ **do**
      $\mathbf{h}^{(0)}_t \leftarrow \mathbf{X}_t$
      **for** $\ell = 1, 2$ **do**
        $e_{t,ij} \leftarrow e^{\mathrm{std}}_{t,ij} + \lambda_\ell \tilde{c}_{t,i}\tilde{c}_{t,j}$ — Eqs. (9)–(10)
        $\mathbf{h}^{(\ell)}_t \leftarrow$ masked-softmax aggregation — Eq. (11)
      $\boldsymbol{\phi}^{(e)}_t \leftarrow \mathrm{FFN}(\mathbf{h}^{(2)}_t)$
    $\mathbf{z}^{\mathrm{route}}_t \leftarrow \sum_{e\in\mathcal{K}_t} \hat{s}_{t,e}\boldsymbol{\phi}^{(e)}_t$ — Eq. (14)
  Encode each node over time; pool; predict $\hat{y}$
  $\mathcal{L} \leftarrow \mathcal{L}_{\mathrm{cls}} + w_{\mathrm{bal}}\mathcal{L}_{\mathrm{bal}}$; Adam step — Eqs. (15)–(16)
**until** early stopping on macro-F1
*Severity estimation (parallel branch)*
$\mathrm{SSI}_v$ from $\tilde{\mathbf{C}}$ — Eqs. (17)–(20)

$\mathcal{L}_{\mathrm{cls}}$, with weight $w_c = |\mathcal{D}^{\mathrm{train}}|/(2n_c)$ for the $n_c$ training clips of class $c$, plus the load-balancing term at weight $w_{\mathrm{bal}} = 0.05$:

$$\mathcal{L} = \mathcal{L}_{\mathrm{cls}} + w_{\mathrm{bal}}\mathcal{L}_{\mathrm{bal}}. \tag{16}$$

Training is staged: CGM initialises every expert of the trained CM-GAT, then learns the gate with Adam (weight decay $10^{-4}$, rate $3 \times 10^{-4}$ for warm-started weights, $10^{-3}$ otherwise), clipping at 1.0, and up to 60 epochs, keeping the checkpoint with the best evaluation-fold macro-F1 (patience 12).

### F. *Stereotypy Severity Index (SSI)*

Beyond binary classification, the scaled tensor $\tilde{\mathbf{C}}$ supports a continuous severity estimate for rehabilitation monitoring, on a branch parallel to the classifier; it takes no input from $\mathbf{z}_{\mathrm{video}}$ or $\hat{y}$, so it is available independently of what the classifier predicts. Following severity measures that score how much, how often, and where stereotyped movement occurs [8], [51], let $\gamma$ denote a fixed chaos-elevation threshold fitted on the reference cohort of all scored clips. For clip $v$, three components are computed over its $T_*$ frames and the node set $\mathcal{J}$: the mean elevation above threshold,

$$X^{\mathrm{int}}_v = \frac{1}{T_*|\mathcal{J}|}\sum_{t=1}^{T_*}\sum_{j\in\mathcal{J}} \max(\tilde{c}_{t,j} - \gamma, 0), \tag{17}$$

the fraction of frames whose node-mean chaos exceeds $\gamma$,

$$X^{\mathrm{temp}}_v = \frac{1}{T_*}\sum_{t=1}^{T_*} \mathbb{1}\left[\frac{1}{|\mathcal{J}|}\sum_{j\in\mathcal{J}} \tilde{c}_{t,j} > \gamma\right], \tag{18}$$

and the fraction of joints whose mean chaos exceeds $\gamma$,

$$X_v^{\text{spat}} = \frac{1}{|J|}\sum_{j\in J} \mathbf{1}\left[\frac{1}{T_*}\sum_{t=1}^{T_*} \tilde{c}_{tj} > \gamma\right], \tag{19}$$

where $\mathbf{1}[\cdot]$ is the indicator function. These separate how far, how often, and how widely irregularity is elevated, three properties a single averaged scalar would conflate. With $\text{pct}(\cdot)$ the 0–100 percentile rank against the reference cohort, the components are averaged with equal weight,

$$\text{SSI}_v = \frac{1}{3}\left[\text{pct}(X_v^{\text{int}}) + \text{pct}(X_v^{\text{temp}}) + \text{pct}(X_v^{\text{spat}})\right]. \tag{20}$$

The complete procedure of the CGM is summarised in Algorithm 1. Chaos formation runs once per clip and has no learnable parameter, so $\tilde{\mathbf{C}}$ is fixed before training and identical at inference; the joint objective of Eq. (16) updates all weights, including every $\lambda_\ell$, and the SSI is read from $\tilde{\mathbf{C}}$ alone.

## III. Experiments and Results

### A. Experimental Settings

All metrics are computed on held-out folds of one fixed subject-grouped five-fold partition (24/23/23/23/23 clips, six TD comparison clips per fold), shared by every configuration, because record-level validation overestimates performance under subject-level dependence [52]. Cohort comparisons use Hedges' $g$ with bootstrap 95 % CIs ($10^4$ resamples) and Holm-corrected Mann–Whitney tests; paired comparisons use the Wilcoxon signed-rank test and are read from the paired effect size $d_z$, since five folds bound its $p$ at 0.0625; multi-group comparisons use the Kruskal–Wallis test. With ASD stereotypy as the positive class, metrics are balanced accuracy (BA), macro-averaged F1 (Macro-F1), sensitivity (Sens.), specificity (Spec.), the Matthews correlation coefficient (MCC), Cohen's $\kappa$, and the areas under the ROC and precision–recall curves (AUROC, AUPRC). Frame-level quantities are reduced to one value per clip before testing, avoiding pseudoreplication.

### B. Mechanistic Validation of Chaos-Modulated Attention

A model may gain accuracy while ignoring the prior it is given, so the mechanism was tested by intervention. Setting $\tilde{\mathbf{C}} = \mathbf{0}$ reduces CM-GAT to standard additive attention; the fold model with the highest evaluation macro-F1 was therefore evaluated twice on identical clips under identical weights, once with the measured tensor (ON) and once with it zeroed (OFF), so every difference in attention is attributable to the modulation term alone (Fig. 4(a)). Attention is averaged over the four experts, two layers, and four heads, then reduced to one paired observation per clip ($N = 116$; 112 for coupling; skeletal edges only).

Although unconstrained, all eight $\lambda_\ell$ remain positive (0.970–0.980); the stronger evidence comes from intervention. Table I reports four per-clip statistics: coupling $\rho$, the Spearman correlation between the attention weight of each skeletal edge and the chaos product $\tilde{c}_i\tilde{c}_j$ of its endpoints; attention entropy, the Shannon entropy of the edge attention weights; Gini concentration (Gini conc.), their inequality; and chaotic-node mass, the share of attention received by the two most irregular joints of each frame. $\Delta$ is the mean ON−OFF difference with bootstrap 95 % CI, Clips[↑] counts clips whose coupling moves in the predicted direction, $N$ is the number of clips, and N/A marks an entry that does not apply. Every paired difference has $p < 0.001$. Coupling rises from $\rho = 0.240$ (OFF) to 0.308 (ON), $d_z = +1.17$, with 105/112 clips (94 %) moving in the predicted direction; the largest change in any single attention weight is 0.0062. The redistribution acts at two levels (Fig. 4(a)): in total mass, attention moves onto self-connections, whose gain exactly cancels the skeletal-edge loss, and the share received by each frame's two most irregular joints rises ($d_z = +1.30$); within the skeletal edges, the remaining weight is re-ranked toward jointly irregular pairs, which the coupling statistic measures. The concentration gain scales with irregularity (Fig. 4(b)): per-clip ΔGini rises monotonically across irregularity quintiles, from $0.67 \times 10^{-3}$ in the least irregular quintile (Q1) to $10.72 \times 10^{-3}$ in the most irregular (Q5). Per-clip paired estimates are given in Supplementary Fig. S1.

TABLE I

Paired Attention Statistics of CGM With the Chaos Tensor Restored (ON) and Zeroed (OFF)

| Statistic | OFF | ON | Δ [95 % CI] | $d_z$ | Clips[↑] | $N$ |
|---|---|---|---|---|---|---|
| Coupling $\rho$ | 0.2401 | 0.3081 | +0.0681 [0.0577, 0.0791] | +1.17 | 105/112 | 112 |
| Attention entropy | 2.5526 | 2.5509 | −0.0017 [−0.0022, −0.0012] | −0.65 | N/A | 116 |
| Gini conc. | 0.1319 | 0.1359 | +0.0040 [0.0031, 0.0050] | +0.79 | N/A | 116 |
| Chaotic-node mass | 0.3106 | 0.3202 | +0.0097 [0.0084, 0.0111] | +1.30 | N/A | 116 |
| Shuffled vs. ON | 0.257 vs. 0.308 | | −0.051 | −1.18 | N/A | 112 |
| Shuffled vs. OFF | 0.257 vs. 0.240 | | +0.017 | +0.43 | N/A | 112 |

To test whether the model needs the tensor's temporal structure or any nonzero signal of the same shape, chaos values were permuted per joint across frames (10 shuffles per clip); the shuffled rows of Table I compare this control with ON and OFF. Coupling under this control ($\rho = 0.257$) falls significantly below the real-chaos condition ($d_z = -1.18$, $p < 0.001$) and significantly above the zeroed baseline ($d_z = +0.43$, $p < 0.001$), recovering roughly one quarter of the ON–OFF gap: most of the coupling increase depends on the temporal pattern of irregularity, not merely its presence.

### C. Expert Routing Analysis

Across all 7,424 routed frames, expert weights shift systematically with frame-level irregularity (Fig. 4(c)): relative to its weight in the low tertile, E2 more than triples (+229 %) and E1 roughly doubles (+97 %) in the high tertile, while E4 declines by over a quarter (−27 %) and E3 stays essentially flat, engaged regardless of regime (Supplementary Table S1). At clip level, each expert's mean routing weight was correlated with clip-level mean irregularity (Spearman, $N = 116$; Fig. 4(d)): E1 ($\rho = +0.446$), E2 ($\rho = +0.552$), and E4 ($\rho = -0.550$) show significant associations in the predicted direction (all $p < 0.001$), while E3 shows none ($\rho = -0.125$, n.s.), consistent with a generalist role. No expert falls below the 0.02 collapse threshold (utilisation 0.046–0.488).

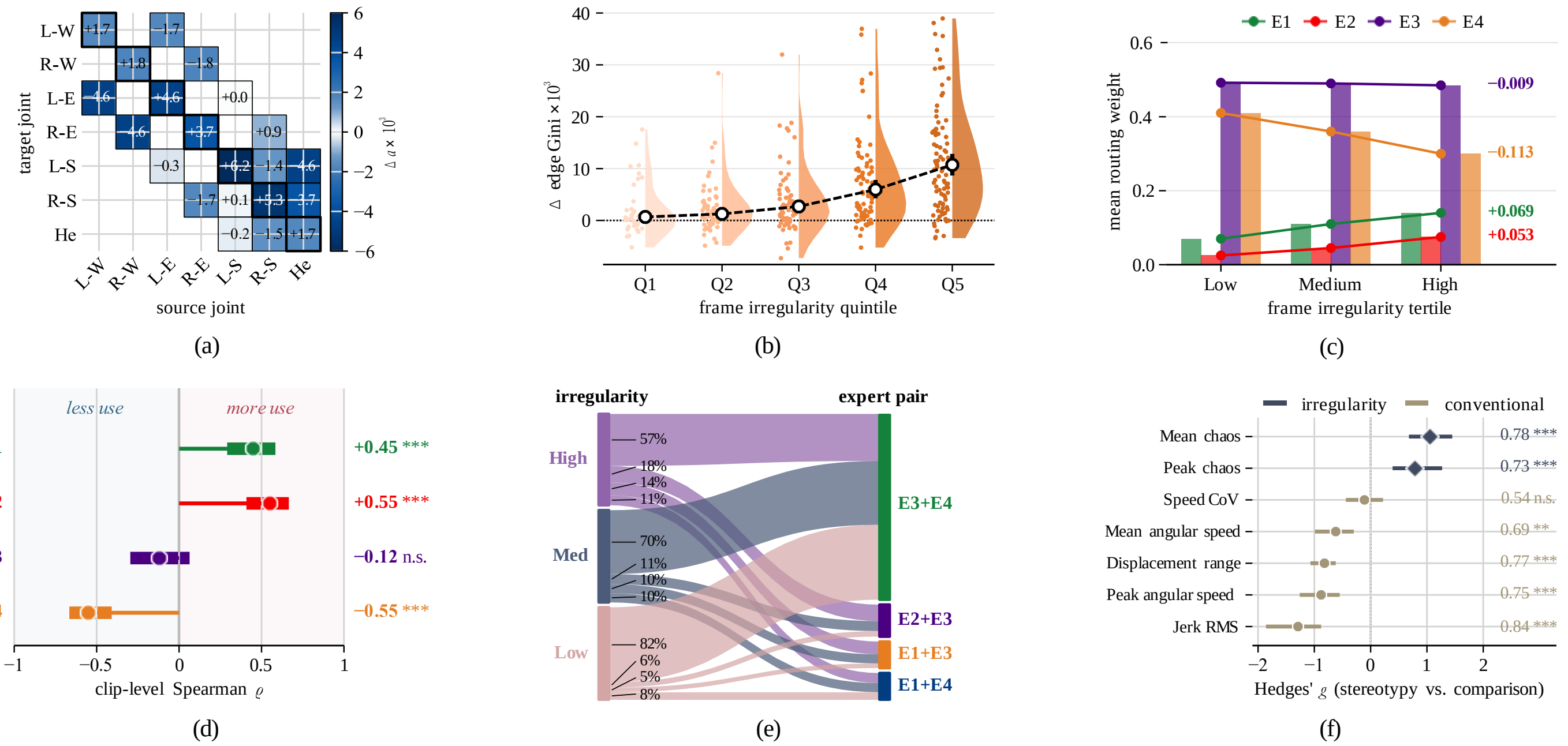

Fig. 4. Mechanistic validation of chaos-modulated attention and chaos-guided routing in the trained CGM. (a) Mean change in attention per joint pair between the restored (ON) and zeroed (OFF) chaos tensor under fixed weights; self-connections outlined. (b) Per-clip change in edge Gini concentration (ON−OFF) across frame-irregularity quintiles; violins show clips, markers the mean with 95 % CI. (c) Mean routing weight of each expert across frame-irregularity tertiles, with the low-to-high change beside each line. (d) Clip-level Spearman correlation between mean irregularity and each expert's routing weight, with 95 % CI (Fisher $z$). (e) Share of frames assigned to each active top-2 expert pair within each irregularity tertile. (f) Hedges' $g$ (ASD stereotypy vs. TD comparison) for irregularity and conventional kinematic descriptors with bootstrap 95 % CI and AUROC; $^{***}p < 0.001$, $^{**}p < 0.01$, n.s. not significant.

At the level of expert pairs (Fig. 4(e)), the dominant pair E3+E4 handles 82 % of low-irregularity frames but only 57 % of high-irregularity frames, the released share moving to pairs containing E1 or E2. The gate also responds to change in irregularity: within each clip, the expert-pair switch indicator correlates with the absolute frame-to-frame change in node-mean irregularity (mean $r = +0.177$ over 116 clips, Wilcoxon $p < 0.001$; Supplementary Section S2).

A gate-input intervention tests whether this association is caused by irregularity input (Fig. 5(a)). With $\langle c \rangle_t$ set to zero (blind gate) or permuted across frames (shuffled gate), and motion input unchanged, the within-clip correlation between each expert's weight and frame irregularity shifts toward zero for all four experts ($|d_z| = 0.75$–$1.56$), most strongly for E4 ($\rho = -0.345 \rightarrow -0.056$, 114/116 clips): routing follows irregularity signal and timing, not only motion content.

## D. *Descriptor Discrimination Analysis*

To justify the choice of signal, we compared mean chaos with five conventional kinematic descriptors by Hedges' $g$ and AUROC (Fig. 4(f)). Mean chaos gives the largest positive effect of any descriptor tested ($g = +1.049$, AUROC $= 0.78$). All five conventional descriptors are negative, and the largest in raw magnitude, jerk RMS ($g = -1.294$, AUROC $= 0.84$), exceeds chaos in $|g|$ but points the opposite way: stereotyped clips show lower jerk than the TD comparison cohort, not higher. Jerk therefore indexes reduced rather than irregular motion. The tensor is also irreducible: reconstructing the full $T \times 7$ tensor from reduced summaries recovers none of its variance from a single clip-level scalar ($R^2 = 0.000$) and at most 30.2 % from a per-frame average, leaving 69.8 % unexplained by marginal summaries (Supplementary Section S3).

## E. *Ablation and Generalisation Performance*

1) ***Component Contribution Analysis:*** Table II compares CGM with four configurations on the SSBD/SSBD+ ASD stereotypy cohort, in order of increasing complexity: Standard GAT, two standard graph-attention layers averaged over time; GAT-MoE, four such experts, initialised from Standard GAT and selected by a gate on motion features alone; Chaos as feature, the chaos value appended to each node's input features; and CM-GAT, chaos-modulated attention without experts. Chaos as feature, CM-GAT, and CGM share one Transformer encoder, so the mechanism comparisons (input versus logit, CM-GAT versus CGM) use matched temporal modelling. The Chaos column states where irregularity enters (input feature or attention logit), the Routing column how experts are selected (motion content or chaos-guided). Gain over Std. GAT is the paired fold-wise difference Δ in BA with bootstrap 95 % CI; Fig. 5 shows each fold (b) and the paired effect size $d_z$ of every gain (c). CGM reaches a BA of 0.930, exceeding standard graph attention by +0.187 with a higher value in all five folds ($d_z = +1.70$) and specificity 1.000 across folds (30/30 TD comparison clips; clip-level Clopper–Pearson 95 % CI [0.884, 1.000]). Its fold-to-fold SD is also the lowest (0.027), under a quarter of the baseline's (0.115). Chaos modulation is the largest single contribution and the only one whose interval excludes zero on its own (Fig. 5(c)). Routing is conditional: added without chaos, its effect is indistinguishable from zero, but added to a chaos-modulated model it is not, because irregularity is the router's informative input, whereas chaos modulation needs no second expert. The point of entry also matters: chaos-as-feature and CM-GAT receive an identical signal at near-identical capacity (228,706 vs. 228,580 parameters), and injecting it at the attention logit outperforms supplying it as an input feature, whose confidence interval includes zero, so the gain reflects where the signal enters, not model size.

TABLE II
ABLATION OF CGM COMPONENTS ON THE PRIMARY COHORT UNDER SUBJECT-GROUPED FIVE-FOLD CROSS-VALIDATION (MEAN ± SD; BEST IN BOLD, SECOND BEST UNDERLINED)

| | Irregularity | | Primary cohort (ASD stereotypy vs. TD comparison) | | | | | | | | Gain over Std. GAT |
|---|---|---|---|---|---|---|---|---|---|---|---|
| Configuration | Chaos | Routing | BA | Macro-F1 | Sens. | Spec. | MCC | $\kappa$ | AUROC | AUPRC | Δ [95 % CI] |
| Standard GAT | N/A | N/A | 0.742 ±0.115 | 0.738 ±0.088 | 0.884 ±0.038 | 0.600 ±0.253 | 0.492 ±0.183 | 0.481 ±0.175 | 0.781 ±0.151 | 0.919 ±0.053 | N/A |
| GAT-MoE | N/A | content | 0.823 ±0.091 | 0.772 ±0.088 | 0.779 ±0.120 | 0.867 ±0.183 | 0.593 ±0.159 | 0.560 ±0.165 | 0.842 ±0.111 | 0.950 ±0.035 | +0.081 [−0.006, +0.169] |
| Chaos as feature | input | N/A | 0.847 ±0.053 | 0.836 ±0.050 | **0.895** ±0.077 | 0.800 ±0.139 | 0.686 ±0.095 | 0.675 ±0.098 | 0.880 ±0.089 | 0.960 ±0.029 | +0.105 [−0.006, +0.176] |
| CM-GAT | logit | N/A | 0.874 ±0.083 | 0.836 ±0.077 | 0.849 ±0.068 | 0.900 ±0.149 | 0.694 ±0.154 | 0.676 ±0.152 | 0.880 ±0.068 | 0.956 ±0.022 | +0.132 [+0.016, +0.226] |
| **CGM (ours)** | logit | chaos-guided | **0.930** ±0.027 | **0.880** ±0.040 | 0.860 ±0.053 | **1.000** ±0.000 | **0.786** ±0.061 | **0.764** ±0.069 | **0.933** ±0.060 | **0.980** ±0.019 | **+0.187** [+0.098, +0.272] |

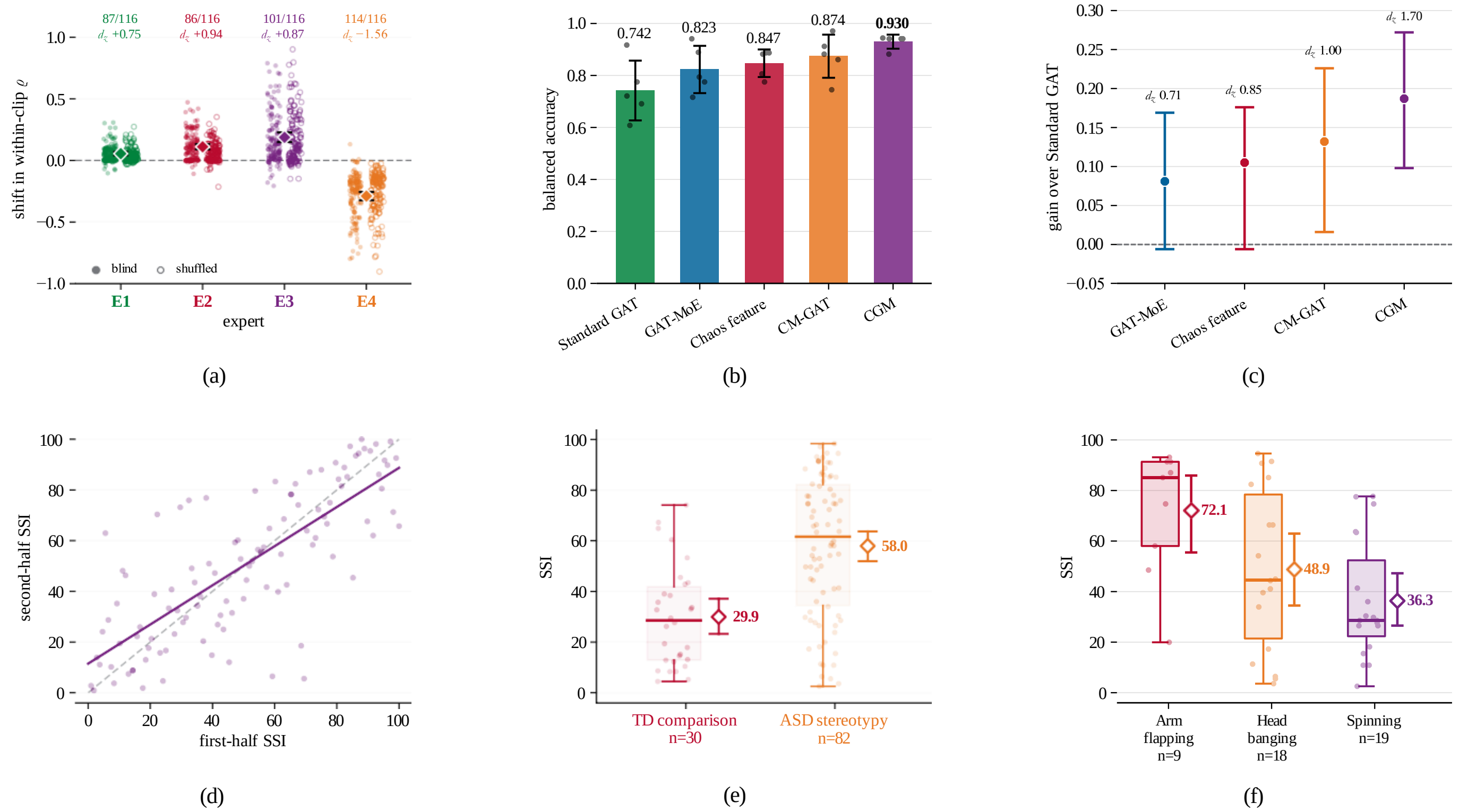

Fig. 5. Gate intervention, classification performance, and validation of the Stereotypy Severity Index (SSI). (a) Per-clip shift in the within-clip Spearman correlation between each expert's routing weight and frame irregularity when the gate's irregularity input is zeroed (filled, blind gate) or frame-shuffled (hollow, shuffled gate); diamonds, mean with bootstrap 95 % CI; labels, clips shifting in the majority direction and $d_z$ . (b) BA of each configuration, mean ± SD over five subject-grouped folds; dots, individual folds. (c) Paired gain in BA over Standard GAT with bootstrap 95 % CI and $d_z$ . (d) Split-half reliability of the SSI; dashed, identity line. (e) SSI of the TD comparison and ASD stereotypy cohorts. (f) SSI across annotated behaviour types. In (e) and (f), boxes show the median and interquartile range, whiskers 1.5× IQR, and diamonds the mean with bootstrap 95 % CI.

*2) **Comparison With Prior Studies:*** Table III positions CGM among recent skeleton- and video-based ASD studies under their published protocols. Because cohorts, tasks, and splits differ, it offers a contextual comparison along two axes: where an auxiliary movement signal enters the network (Signal Entry) and what it encodes (Structural Prior), plus whether a graded severity score is reported. Four studies learn end-to-end from clinical recordings or raw video with no structural prior [8], [38], [42], [43]; among them, Barami *et al.* [8] count stereotyped movements without modelling their temporal structure. Three admit a movement prior only as input content: a joint-angle matrix [25], joint trajectory maps [54], or a self-supervised repetition cue [53] (the last on the same SSBD videos). Zahan *et al.* [25], the closest prior work, shapes node content but not edge weighting. Wang *et al.* [20] is the nearest structural analogue, reshaping adjacency from inter-frame differences, yet encodes displacement rather than irregularity. Skeleton MoE models route by velocity or learned features [29], [30]; to our knowledge, none conditions graph attention or expert routing on measured irregularity, nor derives a severity score from the same representation that drives detection.

TABLE III
POSITIONING CGM AMONG PRIOR SKELETON- AND VIDEO-BASED ASD MOTOR-ANALYSIS STUDIES

| Study | Data | Pose / Joints | Backbone | Signal Entry | Structural Prior | Task | Severity | Reported Result |
|---|---|---|---|---|---|---|---|---|
| [25] (2023) | GFBM lab gait; 59 ASD / 50 TD | Kinect v2 3D, 25 | MS-G3D GCN | Input skeleton | Joint-angle matrix | ASD vs. TD | ADOS regr. ($\rho = 0.34$) | Acc. 93.0±4.8% |
| [8] (2024) | Clinical ADOS videos; 241 ASD | OpenPose 2D, 17 | PoseConv3D | N/A | N/A | SMM detection | SMM count ($r = 0.80$ vs. manual) | Recall 92.5%, Prec. 66.8% |
| [20] (2026) | GFBM; MMASD (32 ASD) | SMPL 3D, 24 | Multi-stream GCN | Graph adjacency | Inter-frame differences | Diagnosis; activity recognition | N/A | GFBM Acc. 90.9%; MMASD Top-1 89.3% |
| [38] (2021) | Clinical ADOS videos; 68 ASD / 68 TD (+101 ASD test) | OpenPose 2D, 18 | VGG16 + LSTM | N/A | N/A | ASD vs. TD | ADOS corr. ($r_s = 0.45$) | Acc. 80.9% |
| [53] (2025) | Extended SSBD home videos | Pose + RGB | Dual video transformer | Self-supervised proxy task | Action repetition | Stereotypy detection | N/A | AUROC 95.0 / 93.1% |
| [43] (2025) | VR lab, 12 tasks; 39 ASD / 42 TD | Azure Kinect 3D, 32 | 3D-CNN ResNet | N/A | N/A | ASD vs. TD | N/A | Acc. 85±3% |
| [42] (2023) | SSBD+ home videos | Raw RGB | CNN pipeline | N/A | N/A | 3 SMM classes + no-class | N/A | Acc. ≈81% |
| [54] (2025) | Clinical paradigms; 121 ASD / 140 TD | HRNet 2D | CNN | Input representation | Joint trajectory maps | ASD vs. TD | N/A | Acc. 87% |
| **CGM (2026)** | **SSBD/SSBD+ (86) vs. Panoptic (30 clips); MMASD (22 ASD)** | **MediaPipe 2D (MMASD: OpenPose 2D), 7** | **GAT experts + sparse MoE + Transformer** | **Attention logit + expert gate** | **Windowed Sample Entropy** | **Stereotypy vs. comparison** | **SSI (split-half 0.87)** | **BA 0.930±0.027; MMASD 0.687** |

*3) **Generalisation Evaluation**:* On MMASD, the full ablation was repeated with subject-grouped five-fold cross-validation and four-class classification, so every test fold contains only children unseen in training (Table IV, upper rows). The two cohorts also test the design under two baseline protocols: on the primary cohort the chaos-free baselines aggregate time by averaging, whereas on MMASD all five configurations share the Transformer encoder of Section II-E. Worst fold is the lowest of the five fold-level BAs, and Folds > Std. GAT counts folds in which a configuration exceeds Standard GAT. Every configuration more than doubles the chance level of 0.25, CGM nearly triples it, and the five-configuration ordering matches that of the primary cohort, so CGM leads under both protocols. CGM exceeds Standard GAT in all five folds ($\Delta = +0.111$, $d_z = +4.14$) and raises the worst-fold BA by +0.122, so its advantage does not rest on favourable subject groups.

*4) **Robustness Evaluation**:* Each configuration was retrained once per fold and its held-out clips were evaluated under seven perturbations applied at inference only (Table IV, lower rows): occlusion of one or two joints, removal of 25 or 50 % of frames, Gaussian jitter of the raw landmarks with SD of 2 or 5 % of shoulder width (SW), and halving of the frame rate (×1/2). For pose noise and frame-rate halving, node features and joint angles were recomputed from the perturbed landmarks, and the change they induce in windowed Sample Entropy was added to the stored tensor. Each cell gives the BA under perturbation (top) and the retention $R$ (bottom), the ratio of that BA to the same model's unperturbed BA; mean BA under perturbation and mean retention (Rob) average the seven conditions. Δ vs. Std. GAT is the paired difference from Standard GAT with its $d_z$. CGM attains the highest BA in five of the seven conditions and the highest mean BA under perturbation (0.798), and exceeds Standard GAT under every perturbation. Landmark jitter leaves it unaffected ($R \geq 98.6$ %), so the irregularity signal is not an artefact of pose-estimation noise. When the frame rate is halved, every configuration that receives the chaos tensor retains a BA of at least 0.729, whereas both chaos-free baselines fall to 0.55; the residual drop of CGM is consistent with windowed Sample Entropy requiring enough samples per window. Within each cohort, the per-clip estimable fraction is uncorrelated with the held-out predicted probability (ASD stereotypy $\rho = -0.087$; TD comparison $\rho = +0.099$; Supplementary Section S4).

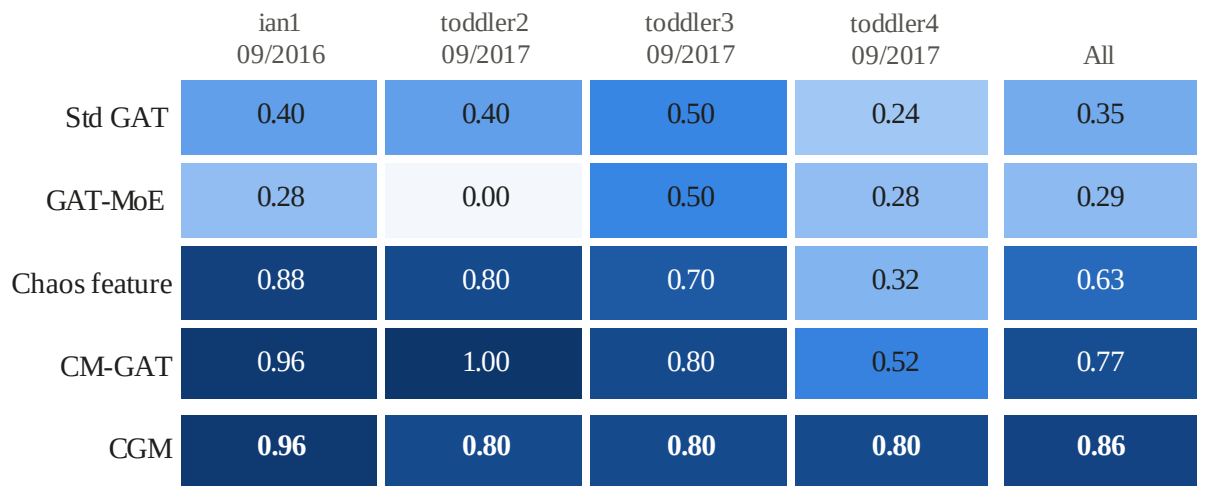

Fig. 6. Held-Out Session Transfer Without Retraining. Share of recordings from four held-out CMU Panoptic Studio sessions classified as TD by each configuration, averaged over the five fold models.

*5) **Held-Out Session Transfer**:* To test whether the comparison class reflects motor behaviour rather than one recording session, every fold model was applied without retraining to four held-out sessions of the Panoptic toddler subset (160906_ian1, 170915_toddler2–4), one captured a year before the training session, through the identical pipeline and Sample Entropy code (Supplementary Section S7). CGM classifies 86 % of the held-out recordings as TD (fold SD 0.06; 92 % for the single model used in the mechanistic analyses) and is the only configuration at or above 0.80 in every session (Fig. 6). Specificity rises from the chaos-free baselines (0.29–0.35) through chaos as feature (0.63) and CM-GAT (0.77) to CGM, so transfer grows with how directly irregularity enters the computation.

TABLE IV
GENERALISATION AND ROBUSTNESS OF CGM AND ITS COMPONENT CONFIGURATIONS ACROSS COHORTS, SUBJECT GROUPS, AND TEST-TIME PERTURBATIONS (BEST IN BOLD, SECOND BEST UNDERLINED)

| Protocol | Level | Standard GAT | GAT-MoE | Chaos as feature | CM-GAT | CGM | Δ vs. Std. GAT ($d_z$) |
|---|---|---|---|---|---|---|---|
| MMASD [7] | Mean BA | 0.576 ±0.090 | 0.590 ±0.145 | 0.595 ±0.149 | 0.667 ±0.149 | **0.687 ±0.088** | +0.111 (+4.14) |
| | Worst fold | 0.409 | 0.313 | 0.377 | 0.371 | **0.531** | +0.122 |
| | Folds > Std. GAT | N/A | 2/5 | 3/5 | 4/5 | **5/5** | N/A |
| Joint occlusion | 1 joint | 0.658 R=89.8% | 0.646 R=85.8% | 0.734 R=91.9% | 0.816 R=93.0% | **0.818** R=91.6% | +0.159 (+1.43) |
| | 2 joints | 0.614 R=83.2% | 0.637 R=84.9% | 0.712 R=88.6% | 0.755 R=86.1% | **0.782** R=87.9% | +0.168 (+1.32) |
| Frame dropout | 25 % | 0.658 R=88.9% | 0.670 R=89.1% | 0.739 R=92.3% | 0.760 R=86.8% | **0.784** R=87.9% | +0.126 (+1.03) |
| | 50 % | 0.647 R=87.1% | 0.653 R=87.3% | **0.717** R=90.5% | 0.636 R=73.2% | 0.703 R=78.7% | +0.056 (+0.34) |
| Pose noise | 2 % SW | 0.742 R=100.0% | 0.751 R=99.7% | 0.790 R=98.2% | 0.872 R=99.7% | **0.890** R=100.1% | +0.148 (+1.29) |
| | 5 % SW | 0.731 R=98.3% | 0.740 R=98.4% | 0.772 R=96.0% | 0.876 R=100.4% | **0.877** R=98.6% | +0.146 (+1.09) |
| Frame-rate halving | ×1/2 | 0.548 R=74.0% | 0.549 R=72.8% | 0.769 R=95.8% | **0.780** R=88.0% | 0.729 R=81.9% | +0.181 (+0.92) |
| Mean BA under perturbation | | 0.657 | 0.664 | 0.747 | 0.785 | **0.798** | +0.141 (+1.09) |
| Mean retention (Rob) | | 88.8% | 88.3% | **93.3%** | 89.6% | 89.6% | N/A |

### F. *Severity Index Reliability Assessment*

The SSI has a split-half reliability of $\rho = 0.77$ (Spearman–Brown-corrected $0.87$; Fig. 5(d)), supporting its use for rehabilitation monitoring. It separates the cohorts (Fig. 5(e)): the TD comparison cohort ($n = 30$) has mean $29.9$ and the ASD stereotypy cohort ($n = 82$; four clips without valid frames excluded) mean $58.0$, a known-groups effect of Hedges' $g = +1.08$ ($p < 0.001$). Across the $46$ clips annotated with a single behaviour, it is sensitive to behaviour type (Fig. 5(f)): arm flapping $n = 9$, mean $72.1$; head banging $n = 18$, mean $48.9$; spinning $n = 19$, mean $36.3$ (Kruskal–Wallis $p = 0.012$). Within-subject consistency holds on MMASD, whose clips come from therapy recordings: for $21$ subjects with $\geq 4$ clips each, split-half across each subject's own clips gives $\rho = +0.775$ ($p < 0.001$). The correlation with an external clinical criterion (ADOS-2 [55] Restricted and Repetitive Behavior subscale, MMASD, $n = 20$) was not significant at this sample size ($\rho = -0.30$, $95\ \%$ CI $[-0.65, +0.17]$). The SSI is therefore a reliable, cohort-referenced motor-axis index that separates the cohorts and holds within subjects across datasets, profiling graded motor irregularity rather than replacing clinical ratings.

## IV. DISCUSSION

Temporal irregularity acts as a structural driver of computation rather than an appended feature: each mechanism reshapes model behaviour on the signal it was designed around, and the accuracy gain follows the point of entry rather than the mere presence of the signal. The principle extends beyond video: brain-connectivity graph models already route or mask by disease-relevant structure [15], [17], [28], [56], and visuomotor [57] and gait-severity [51] assessment could adopt it directly.

Acquisition covaries with group, and the pipeline is built against it through shoulder-based normalisation, joint angles, timestamp-aware differentiation, and masking with $\eta$; every acquisition test of Supplementary Table S4 finds no support for the recording-driven explanation it targets. Appearance cannot enter either: the model sees only seven normalised body joints, no pixels, and facial landmarks, recovered in just $1\ \%$ of comparison frames, are excluded. The TD comparison group is a single participant, so subject grouping holds for the ASD stereotypy cohort but not for the TD comparison class, whose clips enter training and test in every fold. The perfect specificity of Table II is therefore read together with two held-out tests: MMASD, where all $22$ children are split by subject and CGM leads in all five folds, and the held-out Panoptic sessions, where CGM keeps a specificity of $0.86$ without retraining while chaos-free baselines fall to $0.29$–$0.35$ (Fig. 6). Broader typical sampling will further strengthen generalisation [58]. The severity null at $n = 20$ is consistent with underpowering: a comparable skeleton-based effort reached $\rho = 0.34$ ($p = 0.002$) at $n = 61$ [25], and underpowered correlations should not be read either way [59]. Chaos denotes windowed Sample Entropy, following movement-science convention [32], not formal chaos-theoretic dynamics. For rehabilitation, the SSI's stability within children across MMASD therapy recordings is what a session-to-session outcome measure requires, and an ordinary camera suffices to follow stereotypy during home-based intervention [9].

## V. CONCLUSION

This work introduced CGM, a chaos-modulated graph attention network coupled to a chaos-guided sparse mixture of experts for video-only motor screening. CGM reached a balanced accuracy of $0.930$ with the lowest fold-to-fold variability of any configuration, chaos modulation alone contributing $+0.132$; its component ordering replicated on MMASD [7], and its specificity held at $0.86$ on held-out Panoptic sessions without retraining. The same tensor yielded a reliable severity index, so one irregularity representation supports detection, mechanistic interpretation, and graded severity from an ordinary video, a basis for camera-based monitoring of stereotypy during therapy. As the TD comparison cohort is a single participant, multi-participant, multi-site validation, a powered severity study, and a phone-based aid for scalable rehabilitation monitoring are the next steps [58]–[60].